\documentclass{article}
\usepackage{siunitx}
\usepackage{spconf,amsmath,graphicx}
\usepackage{booktabs}
\usepackage{stfloats}
\usepackage{gensymb}
\usepackage{soul}
\usepackage{url}
\usepackage{hyperref}
\usepackage{amssymb}
\usepackage{pifont}
\usepackage[table]{xcolor}

\hypersetup{
    colorlinks=true,
    linkcolor=blue,
    citecolor=blue, 
}
\newcommand{\metric}[1]{\fontsize{7pt}{10pt}\selectfont \textbf{#1}}
\usepackage[commentmarkup=uwave]{changes}
\definechangesauthor[name={Santiago Lopez}, color=red]{slt}

\title{Advancing Limited-Angle CT Reconstruction Through Diffusion-Based Sinogram Completion}
\name{\begin{tabular}{c}
Jiaqi Guo$^{1\dagger}$\thanks{© 20XX IEEE. Personal use of this material is permitted. Permission from IEEE must be obtained for all other uses, in any current or future media, including reprinting/republishing this material for advertising or promotional purposes, creating new collective works, for resale or redistribution to servers or lists, or reuse of any copyrighted component of this work in other works.}, Santiago López-Tapia$^{1}$ and Aggelos K. Katsaggelos$^{1\dagger}$~\thanks{$^{\dagger}$ Corresponding author.}
\end{tabular}}

\address{$^{1}$ Dept. of Electrical and Computer Engineering, Northwestern University, Evanston, IL, USA}

\begin{document}
%
\maketitle
\begin{abstract}
Limited Angle Computed Tomography (LACT) often faces significant challenges due to missing angular information. Unlike previous methods that operate in the image domain, we propose a new method that focuses on sinogram inpainting. We leverage MR-SDEs, a variant of diffusion models that characterize the diffusion process with mean-reverting stochastic differential equations, to fill in missing angular data at the projection level. Furthermore, by combining distillation with constraining the output of the model using the pseudo-inverse of the inpainting matrix, the diffusion process is accelerated and done in a step, enabling efficient and accurate sinogram completion. A subsequent post-processing module back-projects the inpainted sinogram into the image domain and further refines the reconstruction, effectively suppressing artifacts while preserving critical structural details. Quantitative experimental results demonstrate that the proposed method achieves state-of-the-art performance in both perceptual and fidelity quality, offering a promising solution for LACT reconstruction in scientific and clinical applications.
\end{abstract}
\section{Introduction}
\begin{figure*}[htb]\label{structure}
  \centering
  \includegraphics[width=\textwidth]{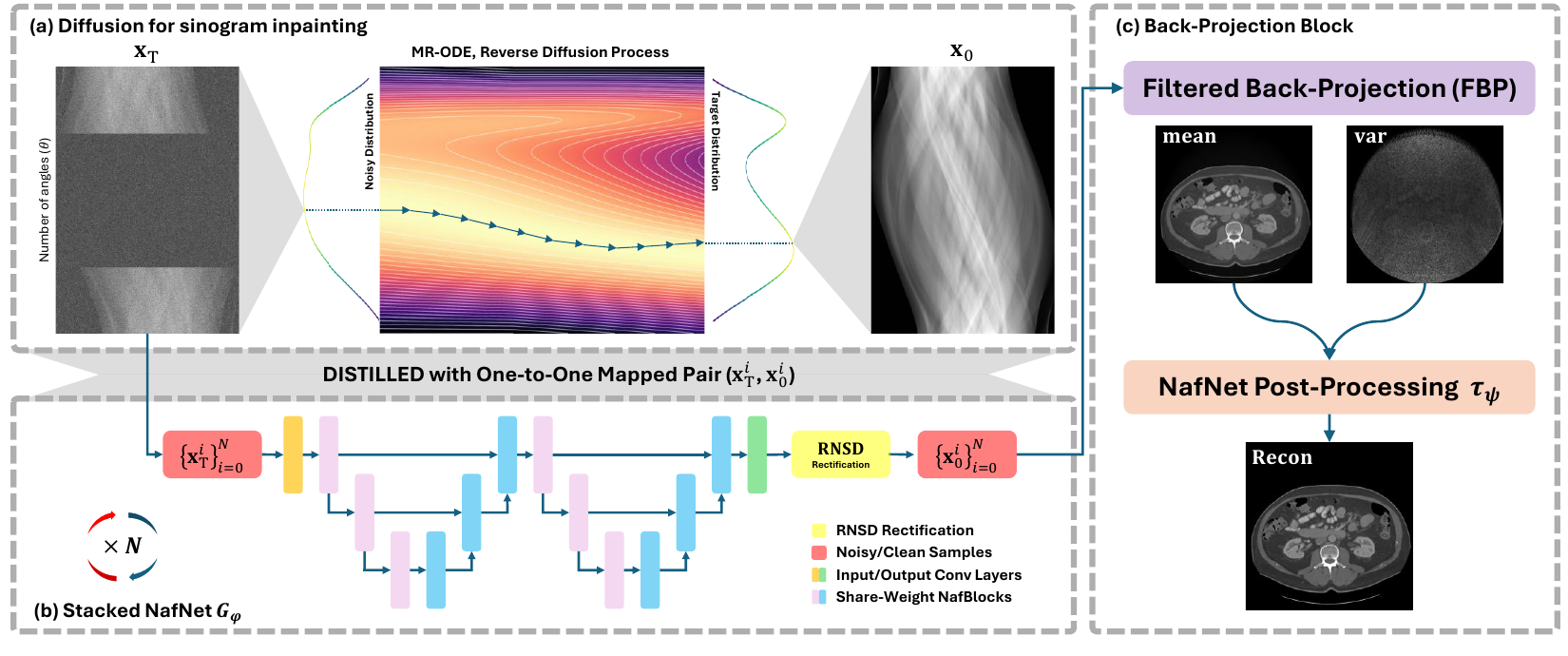}
  \caption{Illustration of our proposed LACT reconstruction pipeline. (a) MR-SDE is employed to restore missing angular information via a diffusion process. (b) Sinogram pairs $(\mathbf{x}_\mathbf{T}, \mathbf{x}_\mathbf{0})$ generated by MR-ODE are used for model distillation. During testing, the distilled model is sampled $N$ times to produce multiple inpainted sinograms $\{\mathbf{x}^i_0\}^N_{i=0}$. (c) The inpainted sinograms are back-projected using FBP, and their mean and standard deviation are computed and fed to the post-processing module $\tau_{\psi}$.
}
\end{figure*}
\label{sec:intro}
Tomographic reconstruction interprets higher-dimensional objects from their projections. Mathematically, this process is described by the Radon Transform~\cite{radon69u}, where projections represent line integrals along scanning trajectories. Ideally, full-angle projections capture complete Fourier domain information, enabling accurate reconstruction. However, constraints in techniques like electron tomography often limit projection angles, leading to the Limited Angle Computed Tomography (LACT) problem, which challenges conventional methods such as Filtered Back Projection (FBP)~\cite{dudgeon1983multidimensional}. In limited-angle computed tomography restoration, the objective is to recover a high-quality image $\mathbf{x}\in\mathbb{R}^{M}$ from a degraded, masked sinogram $\mathbf{y}\in\mathbb{R}^{N}$. This process is typically modeled by the linear equation:
\begin{equation}\label{inverse_problem}
    \mathbf{y} = \mathbf{A}\mathbf{x} + \mathbf{n},
\end{equation}
where $\mathbf{A}\in\mathbb{R}^{N\times M}$ represents the Radon transform that maps an image to its set of projections, and $\mathbf{n}$ denotes white additive noise. Following previous works \cite{liu2023dolce, guo2024rn}, we consider the idealized, noise-free case ($\mathbf{n}=0$).

Traditional deep learning (DL) based Image Restoration (IR) methods usually rely on convolutional neural networks (CNNs) to directly learn the inversion of the Radon transform by minimizing the Mean Squared Error (MSE) between the network output and the ground truth. Although those MSE-based approaches are computationally efficient, it typically yields reconstructions that are overly smooth and lack high-frequency details. The use of DL-based generative models \cite{goodfellow2014generative, kingma2013auto} can significantly enhance the capabilities of DL-based IR methods. Among them, diffusion models~\cite{ho2020denoising} and their variants~\cite{van2017neural, karras2019style, song2019generative, song2020score} have recently beaten Generative Adversarial Networks (GANs)~\cite{dhariwal2021diffusion} and emerged as particularly effective approaches for image restoration. By indirectly modeling the entire image distribution and leveraging a stochastic sampling process, these models are capable of producing high-quality reconstructions that faithfully preserve fine structural details.

Sinogram completion has emerged as an effective strategy for addressing the LACT problem by formulating it as an inpainting task, where missing projection data are inferred from observed measurements. Motivated by the success of DL-based image restoration methods, recent efforts have also explored applying learning-based approaches to address missing information in the sinogram domain. Ghani and Karl~\cite{ghani2018deep} employed a fully convolutional network (FCN) to correct corrupted sinograms by learning a global mapping across the entire sinogram domain, while in a subsequent work, they introduced a conditional GAN architecture for enhanced correction~\cite{ghani2019fast}. Yoo et al.~\cite{yoo2019sinogram} proposed a deep convolutional GAN-based framework further refined with a continuity loss to promote smooth transitions across missing regions. Furthermore, Valat et al.~\cite{valat2023sinogram} incorporated shape priors into a pix2pix-based GAN to regularize the inpainting process. Beyond explicit completion strategies, implicit approaches have also been explored, where networks learn to map incomplete sinograms directly to reconstructed images without explicitly recovering missing projections, as illustrated by Anirudh et al.~\cite{anirudh2018lose}. Although these methods have achieved promising results, they primarily rely on convolutional architectures or adversarial training and have yet to leverage the modeling capabilities of diffusion-based generative frameworks.

Current diffusion-based LACT solutions primarily operate in the image domain, a strategy that optimizes perceptual quality but complicates the enforcement of data consistency. To address this challenge, \cite{liu2023dolce} introduced a proximal optimization step into the reverse diffusion process. However, this method requires multiple forward operations and inverse Radon transforms during each inference step, resulting in prohibitively slow performance. Previously, Guo et al.~\cite{guo2024rn} proposed RN-SDE, a framework that integrates mean-reverting stochastic differential equations (MR-SDE) with Range-Null Space Decomposition (RNSD) based rectification to enforce consistency more effectively. Nevertheless, performing the rectification requires computing the image-domain pseudo-inverse $\mathbf{A}^{\dagger}$ of the Radon transform, which is both a complex and computationally expensive operation. We recognized that performing restoration directly in the sinogram domain transforms the LACT problem into an \textit{inpainting task}, which greatly simplifies the process. In this formulation, the pseudo-inverse is equivalent to the forward operator.

\noindent \textbf{Our Proposed Work}: To provide a more flexible and efficient solution to the LACT problem, the framework proposed in~\cite{guo2024rn} is extended to focus on sinogram inpainting. Specifically, we employ a MR-ODE to complete missing angular information, enhancing reconstruction accuracy while fully preserving the observed information. Moreover, this diffusion model is distilled with a stacked NafNet, accelerating the iterative reverse diffusion process to a single step. Finally, to further suppress artifacts introduced by FBP, we incorporate a post-processing step that refines the back-projected images. Together, these improvements yield more precise and visually consistent CT reconstructions, broadens its applicability across diverse imaging scenarios, and provides a framework for future research.

\section{Method}
\label{sec:Method}

\subsection{MR-SDE for Sinogram Inpainting}
\noindent In this work, we adopt the MR-SDE framework from~\cite{luo2023image} for completing the missing information in the sinogram domain. Similarly, MR-SDE formulates both the forward and reverse processes in a diffusion model as solutions to stochastic differential equations. The forward process, which perturbs a complete sinogram $\mathbf{x}_\mathbf{0}\sim p(\mathbf{x}_\mathbf{0})$ into a noisified version $\mathbf{x}_\mathbf{T}\sim p(\mathbf{x}_\mathbf{T})$ in $\mathbf{T}$ steps, can be written as:
\begin{equation}
\mathrm{d}\mathbf{x}=\zeta_t(\mu-\mathbf{x})\,\mathrm{d}t+\sigma_t\,\mathrm{d}\mathbf{w},
\end{equation}
where $\zeta_t$ and $\sigma_t$ are time-dependent parameters controlling the mean-reversion (drifting) speed and noise level, respectively, and $\mu$ denotes the mean of the terminal state $\mathbf{x}_\mathbf{T}$. The state $\mathbf{x}_t$ can be derived from $\mathbf{x}_\mathbf{0}$ through:
\begin{equation}
p(\mathbf{x}_t|\mathbf{x}_\mathbf{0})=\mathcal{N}\Big(\mathbf{x}_t\mid \mu+(\mathbf{x}_\mathbf{0}-\mu)e^{-\bar{\zeta}_t},\,\lambda^2\big(1-e^{-2\bar{\zeta}_t}\big)\Big),
\end{equation}
where $\zeta_t^{\prime} := \int_{t-1}^t \zeta_z\,\mathrm{d}z$ and $\lambda^2$ is a fixed variance. This process is referred to as "\textit{mean-reverting}" because, as $t\to\infty$, the distribution $p(\mathbf{x}_t)$ converges to $\mathcal{N}(\mu,\lambda^2)$. In our problem settings, the mean $\mu$ corresponds to the masked sinogram, where the masked area corresponds to the missing angular information that needs to be restored.

To recover the complete sinogram from $\mathbf{x}_\mathbf{T}$, the reverse MR-SDE~\cite{luo2023image} is defined as
\begin{equation}\label{reverse_sde}
\mathrm{d}\mathbf{x} = \Big[\zeta_t(\mu - \mathbf{x}) - \sigma_t^2\nabla_{\mathbf{x}}\log p(\mathbf{x}_t)\Big]\,\mathrm{d}t + \sigma_t\,\mathrm{d}\widehat{\mathbf{w}},
\end{equation}
where the score function $\nabla_{\mathbf{x}}\log p(\mathbf{x}_t)$ is estimated by maximizing the likelihood of $p(\mathbf{x}_{t-1}|\mathbf{x}_t,\mathbf{x}_\mathbf{0})$. This is accomplished by employing a neural network $\mathbf{s}_\phi(\mathbf{x}, t)$ to predict the following score:
\begin{equation}
\nabla_\mathbf{x} \log p(\mathbf{x}_t | \mathbf{x}_\mathbf{0})^* = \frac{\mathbf{g}_t\mathbf{x}_t + \mathbf{h}_t\mathbf{x}_\mathbf{0} - (\mathbf{g}_t+\mathbf{h}_t)\mu}{\sigma_t^2 \,\mathrm{d}t}.
\end{equation}
Here, the parameters $\mathbf{g}_t$ and $\mathbf{h}_t$ are defined in terms of $\bar{\zeta}_{t} := \int_0^t \zeta_z\,\mathrm{d}z$ and $\zeta_t^{\prime}$. The derivation is detailed in~\cite{guo2024rn}.

\subsection{Distillation with RNSD-based Rectification}
\label{ssec:ditill}
\noindent 
To accelerate the inference process and further enforce data consistency, we propose training an additional network, $G_\varphi(\cdot)$, to simulate the iterative diffusion process in one step and ultimately enforce data consistency via RNSD-based rectification, as demonstrated in~\cite{guo2024rn}. For simplicity, we adopt the diffusion backbone NafNet~\cite{chen2022simple} as the base structure for the distillation model. Moreover, following the design in~\cite{liu2023instaflow}, we stack two NafNets with shared parameters to achieve improved distillation performance.

Recall that the Wiener process in the reverse diffusion framework permits a nondeterministic mapping from the noisy distribution $p(\mathbf{x}_\mathbf{T})$ to the target distribution $p(\mathbf{x}_\mathbf{0})$, which hinders the distillation of the student model. Fortunately, we can make this mapping deterministic by modifying Equation~\ref{reverse_sde} into its ordinary differential form:
\begin{equation}
\mathrm{d}\mathbf{x} = \Bigl[\zeta_t(\mu - \mathbf{x}) - \frac{1}{2} \sigma_t^2\nabla_{\mathbf{x}}\log p(\mathbf{x}_t)\Bigr]\,\mathrm{d}t,
\end{equation}
such that each noisy input $\mathbf{x}_\mathbf{T}$ yields a unique restoration $\hat{\mathbf{x}}_0$. Note that the model does not need to be retrained, as we have only removed the stochastic component in the SDE. Accordingly, we can train a one-step model via distillation using the following objective:
\begin{equation}~\label{distillation}
\arg\min_{G_\varphi} \Bigl[ \mathbb{D} \Bigl(\mathbf{x}_\mathbf{T} - \hat{\mathbf{x}}_0,\, \hat{G}_\varphi(\mathbf{x}_\mathbf{T})\Bigr) \Bigr].
\end{equation}
Here, $\mathbb{D}(\cdot,\cdot)$ denotes the LPIPS~\cite{zhang2018unreasonable} similarity loss. The variables $\mathbf{x}_\mathbf{T}$ and $\mathbf{x}_\mathbf{0}$ represent the noisified sinogram and its corresponding reconstruction, pre-generated through the MR-ODE. To further enforce data consistency, we introduce an RNSD-based rectification at the end:
\begin{equation}
    \mathbf{x}_\mathbf{0}^{\prime} = \mathbf{A}^{\dagger}\mathbf{y} + \Bigl(\mathbf{I} - \mathbf{A}^{\dagger}\mathbf{A}\Bigr)\hat{G}_\varphi(\mathbf{x}_\mathbf{T}),
\end{equation}
where $\mathbf{y}$ denotes the sinogram with the missing-angle region masked. $\mathbf{A}^\dagger$ represents the pseudo-inverse of the mask operation and is equivalent to $\mathbf{A}$ (this equivalence holds only for inpainting). This refinement can be interpreted as replacing the known angular information (i.e., the null-space content) in $\hat{G}_\varphi(\mathbf{x}_\mathbf{T})$ with the available observation, $\mathbf{A}^\dagger \mathbf{y} = \mathbf{y}$. Experimentally, we found out that this replacement will introduce some disharmony at the boundary. To solve this issue, we add an additional optimization term into Equation~\ref{distillation},
\begin{equation}\label{loss_term2}
\arg\min_{G_\varphi} \Bigl[\mathbb{D}\Bigl((\mathbf{I}-{\mathbf{A}})(\mathbf{x}_\mathbf{T} - \hat{G}_\varphi(\mathbf{x}_\mathbf{T})) + \mathbf{y},\, \mathbf{x}\Bigr)\Bigr].
\end{equation}
During distillation, we use equation~\ref{distillation} and equation~\ref{loss_term2} as a combined objective. We assign a small weighting factor (close to zero) to the optimization term in order to minimize its influence on the distillation process.

For the next step, we employ FBP to project the restored sinogram back into the image domain. An additional NafNet~\cite{chen2022simple} is incorporated for post-processing. The model is optimized using a combination of LPIPS and MSE losses. Notably, we observed that using the mean and variance of the FBP outcomes from multiple reconstructions, i.e., $\{\mathbf{x}^{i}_\mathbf{0}\}^{N}_{i=0}$, as the input to the post-processing network $\tau_\psi$ can remarkably enhance the final restoration quality.

\section{Experiment}
\label{sec:Experiment}
In this section, we present our experimental setup and results. We comprehensively evaluate our proposed method through extensive ablation studies and comparisons with several state-of-the-art approaches. In particular, we investigate three common missing-angle scenarios, 60\degree, 90\degree, and 120\degree, to reflect varying degrees of data incompleteness. To quantitatively assess the restoration quality, we employ three widely recognized metrics: peak signal-to-noise ratio (PSNR) and structural similarity index measure (SSIM) for distortion evaluation, along with the Learned Perceptual Image Patch Similarity (LPIPS)~\cite{zhang2018unreasonable} metric to measure perceptual quality. Finally, we present a qualitative evaluation that includes visual examples of reconstructed CT images and a brief runtime efficiency analysis.

\subsection{General Experiment Setting}
\subsubsection{Dataset and Preprocessing}
We utilized the CT scans from the publicly available 2019 Kidney and Kidney Tumor Segmentation Challenge (C4KC-KiTS) dataset~\cite{heller2019kits19}. This dataset contains approximately 70{,}000 two-dimensional body scans, each of size $512 \times 512$. Since most of these scans are consecutive slices, we perform uniform sampling to obtain around 21{,}100 images. We \textit{randomly selected 24 body scans from different patients for testing} and used the remaining images for training. Following~\cite{heller2021state}, all CT images were clipped to the range of $[-250, 500]$ HU (Hounsfield Units). For the sinogram setting, we adhered to the projection setup of~\cite{guo2024rn}, which yields sinograms of size $720\times512$ corresponding to a uniform scanning interval of $0.25^\circ$. 

\subsubsection{Basic Experiment Setup}
For training the MR-SDE, we followed the configuration in~\cite{guo2024rn} and fixed $\mathbf{T}=200$. In addition to the noisy input, we incorporate a low-fidelity reconstruction to the input. This is obtained from the NafNet reconstruction (see Table~\ref{Tab:compare}, row 3). It is important to emphasize that this additional input is not mandatory and can be replaced by any other form of reconstruction. To generate data for model distillation, we employed the standard DDPM~\cite{ho2020denoising} sampling method to produce 21,000 pairs of $(\mathbf{x}_\mathbf{T},\mathbf{x}_\mathbf{0})$. For each reconstruction, we sample $N=10$ sinograms through the distilled model, back-project them into image domain and compute their mean and standard deviation for subsequent post-processing. All models were implemented with PyTorch and trained using six Nvidia Quadro RTX 8000 GPUs with single-precision ($\texttt{float32}$). We adopted the AdamW optimizer with an initial learning rate of $5\times10^{-4}$, and the learning rate was adjusted using a cosine annealing scheduler. We train the MR-SDE model for 300,000 iterations with a batch size of 8, and perform 100,000 iterations for distillation using a batch size of 16.

\subsection{Ablation Study}
\begin{table}[htb]
\caption{Table of Ablation Studies}\label{tab:fbp_mr_sde}
\centering
\resizebox{0.495\textwidth}{!}{
\setlength{\tabcolsep}{2.5pt}
\begin{tabular}{lccccccccc}
\toprule
\textbf{Angle} &
  \multicolumn{3}{c}{\textbf{60\degree}} &
  \multicolumn{3}{c}{\textbf{90\degree}} &
  \multicolumn{3}{c}{\textbf{120\degree}} \\ \cmidrule(lr){2-4} \cmidrule(lr){5-7} \cmidrule(lr){8-10}
\textbf{Metric} &
  \metric{PSNR$\uparrow$} &
  \metric{SSIM$\uparrow$} &
  \metric{LPIPS$\downarrow$} &
  \metric{PSNR$\uparrow$} &
  \metric{SSIM$\uparrow$} &
  \metric{LPIPS$\downarrow$} &
  \metric{PSNR$\uparrow$} &
  \metric{SSIM$\uparrow$} &
  \metric{LPIPS$\downarrow$} \\ 
\midrule
\multicolumn{8}{l}{\textit{Ablation on Diffusion Distillation (w/o Post-Processing)}} \rule{0pt}{1.5ex}\\
\midrule
\textbf{w/ distill}
& 31.20 & 0.740 & 0.0808
& 30.21 & 0.707 & 0.0972
& 28.57 & 0.674 & 0.1226 \\

\textbf{w/o distill} 
& 30.36 & 0.718 & 0.0948
& 28.92 & 0.665 & 0.1321
& 27.34 & 0.598 & 0.1748 \\ 

\midrule
\multicolumn{6}{l}{\textit{Ablation on Post-Processing}} \rule{0pt}{1.5ex}\\
\midrule
\textbf{w/o $\mathcal{L}_{lpips}$}
& 37.82 & 0.972 & 0.0509
& 33.68 & 0.949 & 0.0780
& 30.51 & 0.919 & 0.1022 \\

\textbf{w/o sinoinp} 
& 36.57 & 0.966 & 0.0135
& 33.14 & 0.941 & 0.0247
& 29.77 & 0.906 & 0.0413 \\ 

\bottomrule
\end{tabular}
\hspace{+10pt}}
\end{table}\vspace{-10pt}
Table~\ref{tab:fbp_mr_sde} presents the results of our ablation studies. First, we compare the performance of models that use diffusion distillation for sinogram inpainting (\textbf{w/ distill}) against models that directly minimize the mean squared error objective without distillation (\textbf{w/o distill}). To ensure a fair comparison, all evaluations are based on the FBP outcomes across three distinct LACT scenarios. The results clearly demonstrate that incorporating diffusion distillation yields superior performance, as evidenced by higher PSNR and SSIM values and lower LPIPS scores across all angular settings.

Next, we assess the impact of two factors in the post-processing stage. The configuration labeled \textbf{w/o $\mathcal{L}_{lpips}$} omits the LPIPS perceptual loss during the training of the post-processing block, while the configuration labeled \textbf{w/o sinoinp} bypasses sinogram inpainting entirely by directly applying the post-processing block to the FBP outcomes in the image domain. Comparing our results in Table~\ref{Tab:compare}, we observe notable performance degradation in both cases—particularly in terms of perceptual quality.

\subsection{Evaluation and Comparison}
\begin{table}[htbp]\vspace{-10pt}
\caption{Comparison of Different Methods on C4KC-KiTs}
\label{Tab:compare}
\centering
\resizebox{0.495\textwidth}{!}{
\setlength{\tabcolsep}{2.5pt}
\begin{tabular}{lccccccccc}
\toprule
\textbf{Angle} & \multicolumn{3}{c}{\textbf{60°}} & \multicolumn{3}{c}{\textbf{90°}} & \multicolumn{3}{c}{\textbf{120°}} \\
\cmidrule(lr){2-4} \cmidrule(lr){5-7} \cmidrule(lr){8-10}
\textbf{Metric} & \textbf{PSNR↑} & \textbf{SSIM↑} & \textbf{LPIPS↓} & \textbf{PSNR↑} & \textbf{SSIM↑} & \textbf{LPIPS↓} & \textbf{PSNR↑} & \textbf{SSIM↑} & \textbf{LPIPS↓} \\
\midrule
\textbf{FBP}      
& 16.22 & 0.316 & 0.1914
& 13.92 & 0.255 & 0.2771
& 11.34 & 0.178 & 0.3688 \\
\textbf{TV}~\cite{chambolle2011first} 
& 23.37 & 0.489 & 0.2959
& 21.37 & 0.379 & 0.3380
& 19.48 & 0.330 & 0.3775 \\\midrule
\textbf{NafNet}~\cite{chen2022simple}       
& 37.57 & 0.971 & 0.0518
& 33.68 & 0.949 & 0.0776
& 30.29 & 0.918 & 0.1037 \\\midrule
\textbf{DOLCE}~\cite{liu2023dolce} 
& 35.80 & 0.956 & 0.0180
& 32.41 & 0.928 &  0.0325
& 28.57 & 0.879 &  0.0479 \\
\textbf{RN-SDE}~\cite{guo2024rn}  
& 36.86 & 0.960 & 0.0257
& 32.89 & 0.936 & 0.0393
& 29.53 & 0.902 & 0.0629  \\\midrule
\textbf{Ours}
& 37.51 & 0.968 & 0.0118
& 33.91 & 0.944 & 0.0222
& 30.29 & 0.912 & 0.0375 \\
\bottomrule
\end{tabular}
\hspace{+10pt}}
\end{table}
In this section, we compare our proposed method with several state-of-the-art LACT reconstruction techniques. Table~\ref{Tab:compare} presents the ten-run average results for PSNR, SSIM, and LPIPS across three angular settings (60\degree, 90\degree and 120\degree). The evaluated methods include classical approaches (FBP and TV~\cite{chambolle2011first}), MMSE-based IR methods (NafNet~\cite{chen2022simple}), and diffusion-based IR methods (DOLCE~\cite{liu2023dolce} and RN-SDE~\cite{guo2024rn}). It is important to note that the results for NafNet, DOLCE, and RN-SDE are directly taken from~\cite{guo2024rn}, and these methods all operate in the image domain. 

As expected, compared with diffusion-based methods such as DOLCE and RN-SDE, our approach outperforms them on distortion metrics (PSNR and SSIM), achieving an improvement of approximately 1 dB in PSNR while also leading in perceptual quality (LPIPS). When compared with NafNet, and taking into account the perceptual-distortion tradeoff~\cite{blau2018perception}, our method nearly closes the gap in distortion metrics while demonstrating a substantial advantage in perceptual quality.

\subsection{Visual Evaluation and Runtime Efficiency}
\label{ssec:visual}

\begin{figure}[htb]
  \centering
  \includegraphics[width=0.495\textwidth]{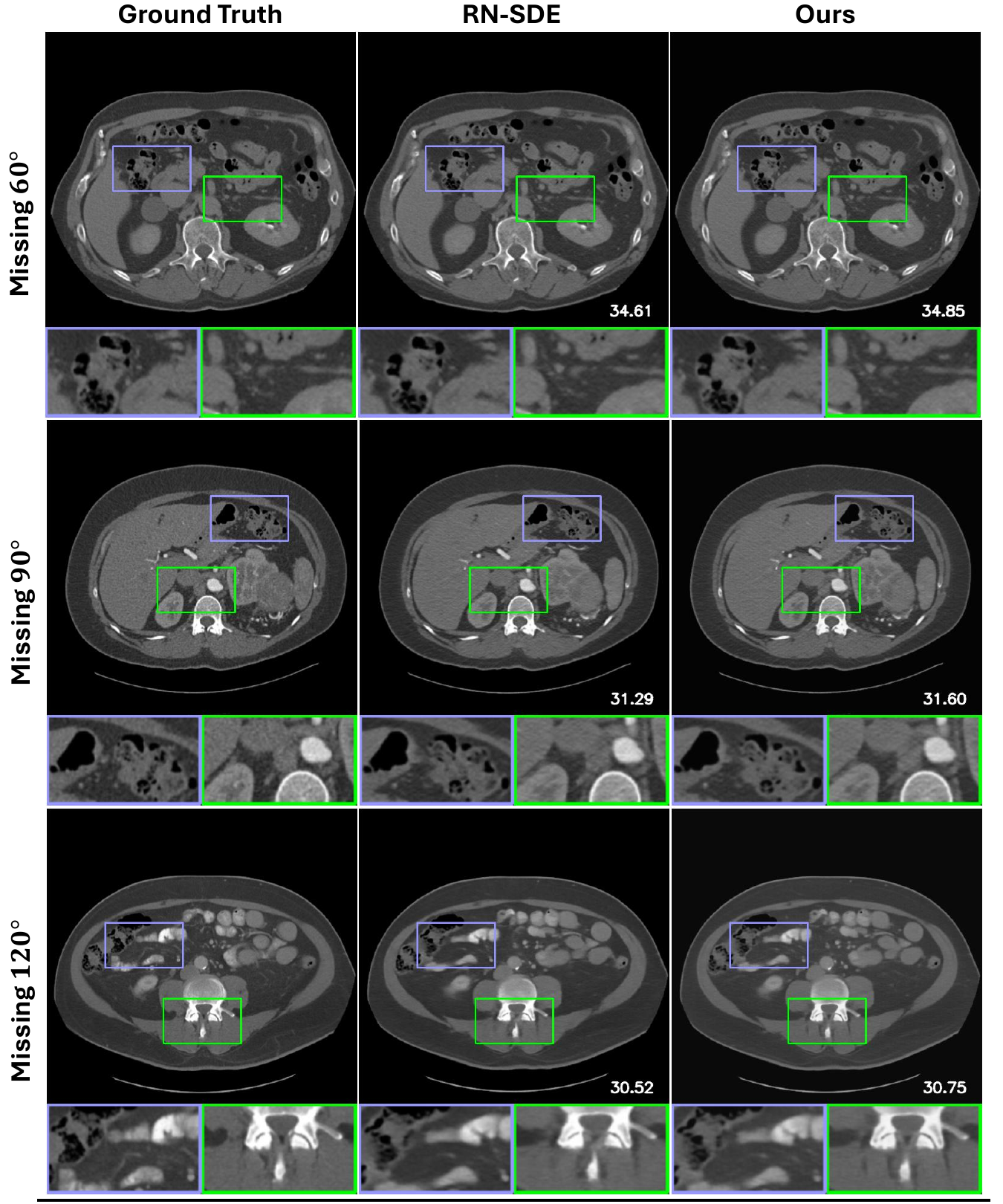}
  \caption{Visualization of reconstructions generated by RN-SDE and our proposed method under three LACT scenarios. The number at the bottom right corner shows the corresponding PSNR relative to the Ground Truth. (\textbf{first column})}
  \label{fig:visualization}\hspace{+5pt}\vspace{-10pt}
\end{figure}
We provide restoration examples under three different missing-angle scenarios, as shown in Figure~\ref{fig:visualization}. In terms of both perceptual quality and data consistency, the differences between RN-SDE and our method are nearly imperceptible. However, when considering time consumption, employing distillation enables our approach to complete the restoration process nearly instantaneously. In Table~\ref{Tab:run_time} shows the runtime efficiency of our method compared against two other diffusion-based methods. For each test, we maintained a batch size of 1, used identical input data, and recorded the runtime. It is evident that our proposed one-step method is significantly faster than the other two diffusion-based approaches (almost a $50\times$ and a $5000\times$ speed up with respect of RN-SDE~\cite{guo2024rn} and DOLCE~\cite{liu2023dolce}, respectively).

\begin{table}[htbp]\vspace{-10pt}
\caption{Comparison of Runtime Efficiency, Experiments are based on a single Nvidia Quadro RTX 8000 GPU, and an Intel(R) Xeon(R) Gold 6226R CPU}\label{Tab:run_time}\vspace{+2pt}
\centering
\resizebox{0.495\textwidth}{!}{
\setlength{\tabcolsep}{2.5pt}
\begin{tabular}{lccccc}
\toprule
\textbf{Method} \rule{0pt}{2.0ex} & \textbf{Backbone} & \textbf{Sino Domain} & \textbf{Img Domain} & \textbf{Time} & \textbf{Iters}\\ 
\midrule
\textbf{DOLCE~\cite{liu2023dolce}} & CondUNet & \ding{55} & \checkmark & 10875s & 2000\\
\textbf{RN-SDE~\cite{guo2024rn}} & CondNafNet & \ding{55} & \checkmark & 106s & 585\\
\textbf{Ours} & StackedNafNet & \checkmark & \ding{55} & 2.33s & 1\\
\bottomrule
\end{tabular}
\hspace{5pt}}\vspace{1pt}
\\
\raggedright
{\hspace{5pt}\scriptsize \parbox{\dimexpr\linewidth-8pt}{Note: Different implementations may result in variations in code execution efficiency, the times presented are approximate and intended for reference only.}}
\end{table}

\section{Conclusion \& Discussion} \label{sec:foot}
In this work, we presented a novel framework for limited-angle computed tomography (LACT) reconstruction that leverages sinogram inpainting via MR-SDE and one-step distillation. Our method first addresses the missing angular information in the sinogram domain. In the distillation stage, we incorporate an RNSD-based rectification to enforce data fidelity effectively. Subsequently, filtered back-projection (FBP) followed by a dedicated post-processing module serves a dual purpose: it projects the inpainted sinogram back into the image domain and further refines the reconstruction.

Comprehensive experiments on the C4KC-KiTS dataset, along with extensive ablation studies, confirm the efficacy of our approach. Notably, our method achieves significant improvements in distortion metrics—demonstrating approximately a 1 dB gain in PSNR over diffusion-based methods—while also yielding superior perceptual quality. When compared with MMSE-based methods like NafNet, our approach nearly closes the gap in distortion metrics. These results highlight the effectiveness of our sinogram inpainting strategy combined with rigorous data fidelity enforcement via RNSD-based rectification, and they underscore the importance of balancing the perceptual-distortion tradeoff in LACT reconstruction. Furthermore, our diffusion distillation technique accelerates the inference process by reducing the iterative procedure to a single-step model, without compromising reconstruction quality. This efficiency, combined with the improved image quality, positions our method as a promising solution for both scientific and clinical applications where rapid and accurate CT reconstructions are critical.

\bibliographystyle{IEEEbib}

\bibliography{strings,refs}

\end{document}